%% file: main.tex
\PassOptionsToPackage{table, prologue}{xcolor}

\pdfoutput=1
\documentclass[11pt]{article}

\usepackage[]{emnlp2022}

\usepackage{times}
\usepackage{latexsym}
\usepackage{caption}
\usepackage{subcaption}
\usepackage{multirow}
\usepackage{textcomp}  % Required for encoding \textbigcircle
\usepackage{scalerel}  % Required for emoji \scalerel

\usepackage[T1]{fontenc}
\usepackage[utf8]{inputenc}

\usepackage{microtype}
\usepackage{pifont}
\newcommand{\cmark}{\ding{51}}%

\usepackage{amsmath,amssymb}
\usepackage{comment}
\usepackage[export]{adjustbox}

\definecolor{lightgray}{gray}{0.9}

\usepackage{xspace}
\usepackage{booktabs}
\usepackage{multicol}
\usepackage{amsmath}
\usepackage{multirow}
\usepackage{graphicx}
\usepackage{enumitem}
\usepackage{mdwlist}
\usepackage{textcomp}
\usepackage{siunitx}
\usepackage{longtable}
\usepackage{pbox}
\usepackage{amssymb}
\usepackage{xspace}

\newcommand{\sref}[1]{\S\ref{#1}}
\newcommand{\fref}[1]{Figure~\ref{#1}}
\newcommand{\tref}[1]{Table~\ref{#1}}

\renewcommand*{\paragraph}{\noindent\textbf}

\newcommand{\potato}[0]{\textsc{Potato}\xspace}

\title{\textsc{Potato}: The Portable Text Annotation Tool}

\author{
 Jiaxin Pei $^{1}$ ~ Aparna Ananthasubramaniam$^{1}$ ~ Xingyao Wang$^{2}$ ~ Naitian Zhou $^{3}$ ~  \\ \textbf{Apostolos Dedeloudis}$^{4}$
 \textbf{Jackson Sargent}$^{1}$ ~ \textbf{David Jurgens}$^{1}$ \vspace{0.1cm} \\
$^1$School of Information, University of Michigan, Ann Arbor, MI, USA  \\
$^2$Department of Computer Science, University of Illinois, Urbana Champaign, IL, USA\\
$^3$School of Information, University of California, Berkeley, CA, USA\\
$^4$The American College of Greece, Athens, Greece\\
{ \tt $^{1}$\{pedropei, akananth, jacsarge, jurgens\}@umich.edu}\\
{  \tt $^{2}$xingyao6@illinois.edu} {  \tt $^{3}$naitian@berkeley.edu}
{  \tt $^{4}$apostolosded@gmail.com}\\
}

\date{}

\begin{document}
\maketitle
\begin{abstract}

We present \potato, the \textbf{Po}rtable \textbf{t}ext \textbf{a}nnotation \textbf{to}ol, a free, fully open-sourced annotation system that 1) supports labeling many types of text and multimodal data; 2) offers easy-to-configure features to maximize the productivity of both deployers and annotators (convenient templates for common ML/NLP tasks, active learning, keypress shortcuts, keyword highlights, tooltips); and 3) supports a high degree of customization (editable UI, inserting pre-screening questions, attention and qualification tests).
Experiments over two annotation tasks suggest that \potato improves labeling speed through its specially-designed productivity features, especially for long documents and complex tasks. \potato is available at {\small \url{https://github.com/davidjurgens/potato}} and will continue to be updated. %A video introduction is available at {\small \url{https://youtu.be/L8A8SIPLBR4}}.
%\daj{someone please write an abstract}.

\end{abstract}

\section{Introduction}

Much of NLP requires annotated data. As NLP has tried to tackle increasingly more complex linguistic phenomena or diverse labeling and classification tasks, the annotation process has increased in complexity---yet the need for and benefit of large labeled datasets remain \cite{halevy2009unreasonable,sun2017revisiting}. Modern annotation tools like Label Studio \cite{tkachenkolabel}, LightTag \cite{perry2021lighttag}, Doccano \cite{doccano}, and Prodigy \cite{prodigy2017} have partially filled this gap, providing a variety of solutions to different types of annotations. However, these tools each bring their own challenges: requiring external access, limiting visual configurability for complex tasks, or even costing hundreds of dollars---prohibitive for small groups. We introduce \potato, The  \textbf{Po}rtable \textbf{t}ext \textbf{a}nnotation \textbf{to}ol, which allows practitioners to quickly design and deploy complex annotations tasks.

\begin{figure}[t]
% \centering
\includegraphics[width=0.49 \textwidth]{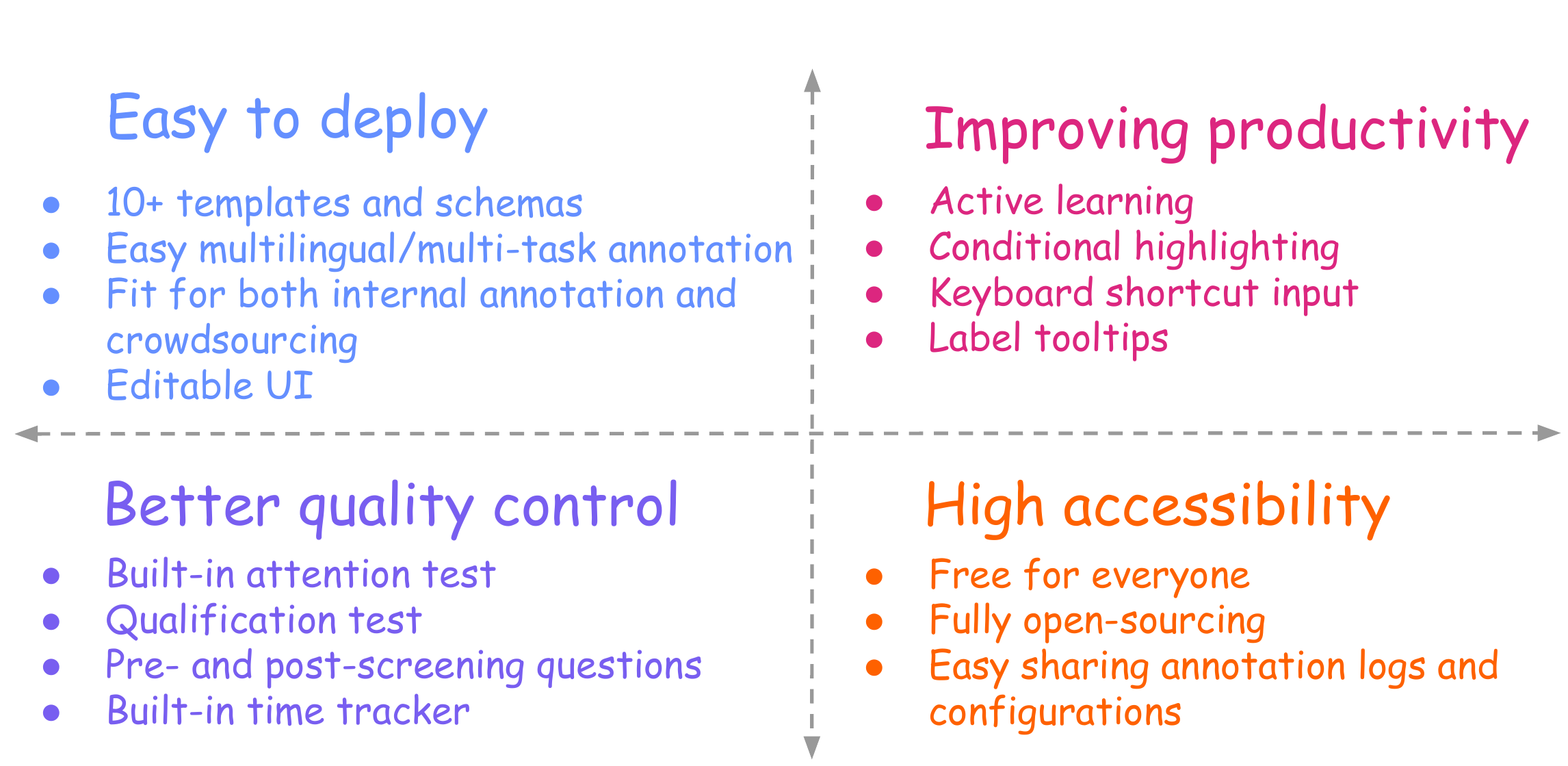}
\caption{The four core design goals of \potato: easy to deploy, greater productivity, better quality control, and high accessibility. Each design goal comes with a series of features that can make data annotation easier and more reliable.}
\label{fig:portable-meaning}
\end{figure}

\potato has been designed, developed, and tested over a two-year period with the following design goals in mind (Figure \ref{fig:portable-meaning}):
1) \textbf{High accessibility}. \potato is open-sourced under the MIT license and free to anyone. \potato is built with minimal dependencies to allow researchers and developers to easily build and integrate their own features. 
2) \textbf{Easy to deploy}. \potato comes with templates covering a wide range of annotation tasks like best-worst-scaling, text classification, and multi-modal conversation. Anyone can start a new annotation project with simple configurations. \potato is also rapidly and easily deployable in local and web-based configurations and has seamless integration with common crowdsourcing platforms like Amazon Mechanical Turk and Prolific. \potato flexibly supports diverse annotation needs. With our specially designed schema rendering and custom rendering mechanism, \potato allows nearly all kinds of text annotation tasks and is visually customizable to support complex task designs and layout. 
3) \textbf{Better quality control}. Attaining reliable annotations is one of the core goals of data annotation tasks. \potato is designed with a series of features that can help to improve annotation quality, including built-in attention tests, prestudy qualification tests, and an annotation time tracker. \potato also allows deployers to easily set up pre- and post-screening questions (e.g. demographics or psychological surveys), which can help researchers to better understand potential biases in data labeling. 
4) \textbf{Productivity enhancing}. \potato comes with a series of productivity features for both deployers and annotators like active learning, conditional highlighting, and keyboard shortcuts. 
%
%5) 
%
While existing systems like Doccano \citep{doccano} and Lighttag \citep{perry2021lighttag} offer different subsets of these features, \potato aims to support a holistic annotation experience by meeting all of these needs. Experiments on two annotation tasks demonstrate that \potato leads to more efficient data labeling for complex tasks.

\section{Architecture and Design}

% Data module, 
% annotation schema/front end
% agreement management 
% active learning

\begin{figure*}[t]
\centering
\includegraphics[width=0.95\textwidth]{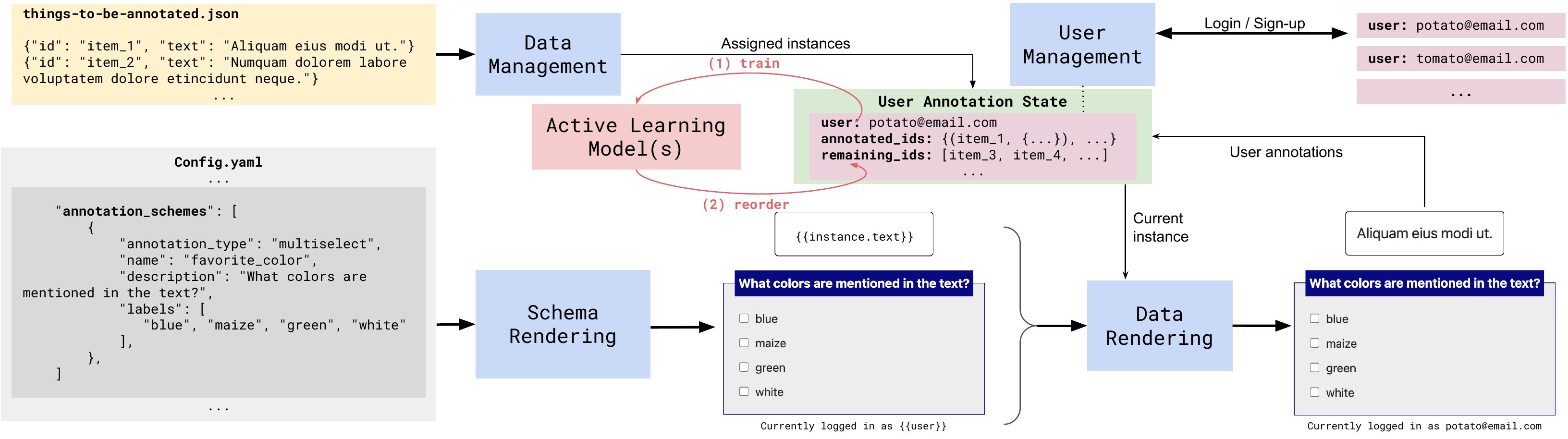}
\caption{The overall architecture of \potato features a modular design that decouples the  task specification from the rendering, allowing rapid deployment of new task designs and separate customization of the visualization. }
%
%\daj{Add more explanation here of what a reader should learn from this figure}}
\label{fig:architecture}
\end{figure*}

\potato is  written in Python and focuses on portability and simplicity in annotation and deployment. The user interface is created through an extensible HTML template and configuration file, which allows practitioners to quickly develop and deploy common setups like Likert-scale annotation while also supporting extensive display customization. The \potato server populates the interface with data provided by the operator and supports displaying any HTML-supported modality, e.g., text or images.
%The server handles  multi-user annotation, annotator agreement management, and active learning for efficient annotation. 
An overview of the architecture is shown in \fref{fig:architecture}.

\paragraph{Data Management}
\potato loads data in common file formats, such as delimited files or newline-delimited JSON. This allows it to ingest data in the JSON format supplied by the Twitter or Reddit APIs, as well as other types of data used by the deployer. All formats are converted into internal data structures that link the deployer-selected instance ID to annotations.  At a minimum, deployers must specify which field represents the unique instance ID and, for most tasks, the text to be annotated. The data may contain other columns which will be included in the final output and can optionally be used in customized visualizations. %The data management module also keeps track of the assignment and annotation status of each instance. 

\paragraph{Annotation Schema Rendering}
\potato allows deployers to select one or more forms of annotation for their data using predefined schema types, such as multiple choice or best-worst scaling. Deployers fill out which options should be shown and then each scheme is rendered into  HTML upon the completion of loading the data. 
Annotation instructions can be provided as an external URL that annotators may view or using HTML text shown in \potato that annotators may collapse vertically to free up screen space.
\potato provides default HTML templates that automatically lay out each scheme's annotation questions. However, deployers may additionally customize the HTML templates and select their own layout using JINJA expressions \citep[e.g., \texttt{\{\{text\}\}};][]{jinja} to specify where parts of the annotation task and data are populated within the user-defined template.

\paragraph{User Management}
Annotators create accounts and then log in to view their tasks using a secure user management system. When used with crowdsourcing platforms, \potato also allows workers to directly jump to the annotation task using their crowdsourcing user ID. %\footnote{For example, \url{annotation-site.com/PROLIFIC_ID=test_user} automatically directs the user to the annotation page without the signup procedure} 
For each new annotator, \potato automatically assigns instances as configured by the deployer and all the annotations are recorded on the backend. When logging out and back in, annotators resume at the most recent unannotated item. \potato also allows deployers to, with minimum configurations, set up pre- and post-screening questions (e.g., having annotators provide demographics or complete psychological questionnaires), pre-study tests, and attention tests to identify unreliable annotators.

\paragraph{Active Learning}
In  some settings such as data with imbalanced classes, active learning helps re-order items to surface those that may provide more information to downstream classifiers \cite{settles2009active,monarch2021human}. Prior annotation interfaces have included active learning to help maximize data utility  \cite{stenetorp2012brat,wiechmann2021activeanno,li2021fitannotator}. \potato includes a configurable active learning setup to prioritize important samples and potentially improves data quality with limited labeling budgets. In its default setting, \potato periodically trains a logistic regression classifier using unigram and bigram features on the currently annotated data; unlabeled instances are sorted by classifier confidence and items with low confidence are prioritized, while still including a deployer-specified percentage of a random sample. Deployers may change or reconfigure this model easily.

\paragraph{Design highlights} \potato is designed to flexibly support diverse annotation tasks and improve the productivity of annotators. Here we briefly highlight several features of \potato. First, with simple configurations, deployers can quickly add \textit{keyboard shortcuts} to specific options or \textit{tooltips} to help annotators. Second, in settings where an annotator is reading a dense or long passage, or where there are many potentially subtle cues, annotators are likely to struggle  due to having to slowly and carefully read each passage or  accidentally omitting relevant annotations due to the complexity of the task. \potato introduces a new feature, \textit{conditional highlighting}, to help in these settings, where the deployer specifies certain keywords to trigger highlights in the text, drawing the annotator's focus to those words or phrases. 
For example, if annotating for Twitter-based stance towards politicians, a deployer might use keywords and phrases for common politicians or political parties to ensure these are not missed. 
If conditional highlighting is enabled, \potato will also randomly label some words with highlights, based on a deployer-specified rate, to ensure annotators do not rely too heavily on highlights.

\begin{figure*}[t]
    \centering
    \begin{subfigure}[b]{0.3\textwidth}
        \centering
        \includegraphics[width=\textwidth]{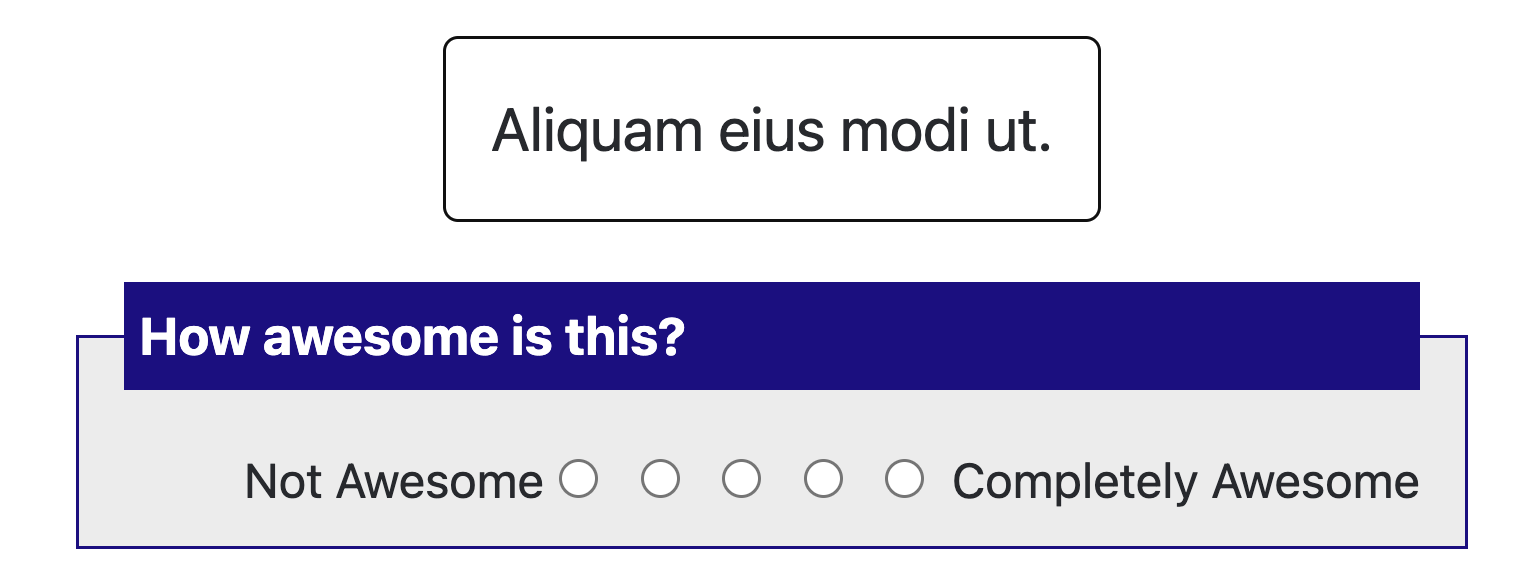}
        \caption{Likert Ratings}    
        \label{fig:likert}
    \end{subfigure}
    \hfill
    \begin{subfigure}[b]{0.2\textwidth}  
        \centering 
        \includegraphics[width=\textwidth]{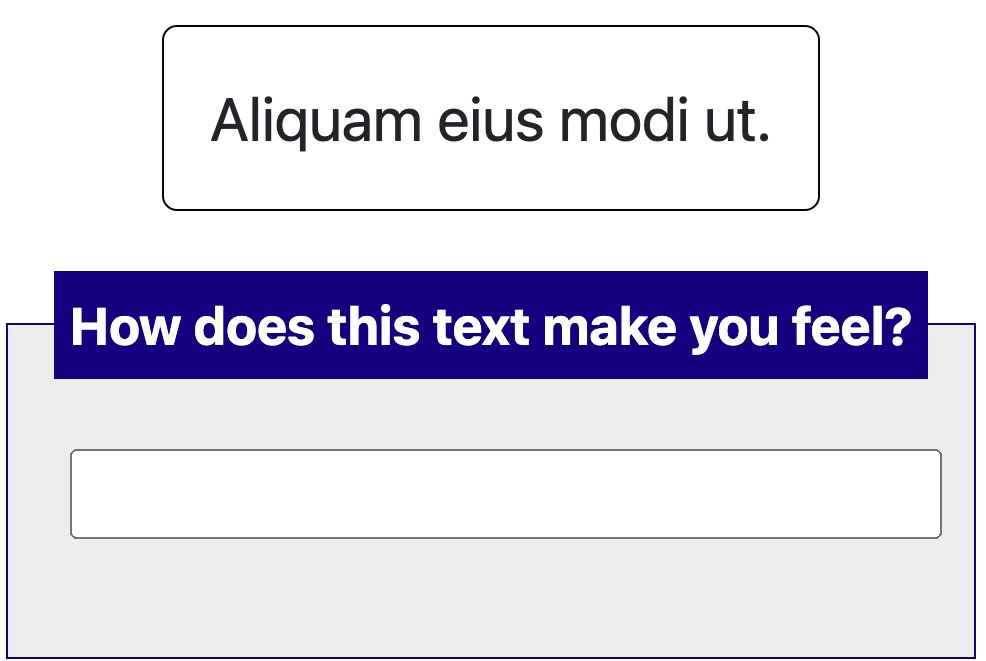}
        \caption{Text-box}    
        \label{fig:textbos}
    \end{subfigure}
    %\vskip\baselineskip
    \hfill
    \begin{subfigure}[b]{0.4\textwidth}   
        \centering 
        \includegraphics[width=\textwidth]{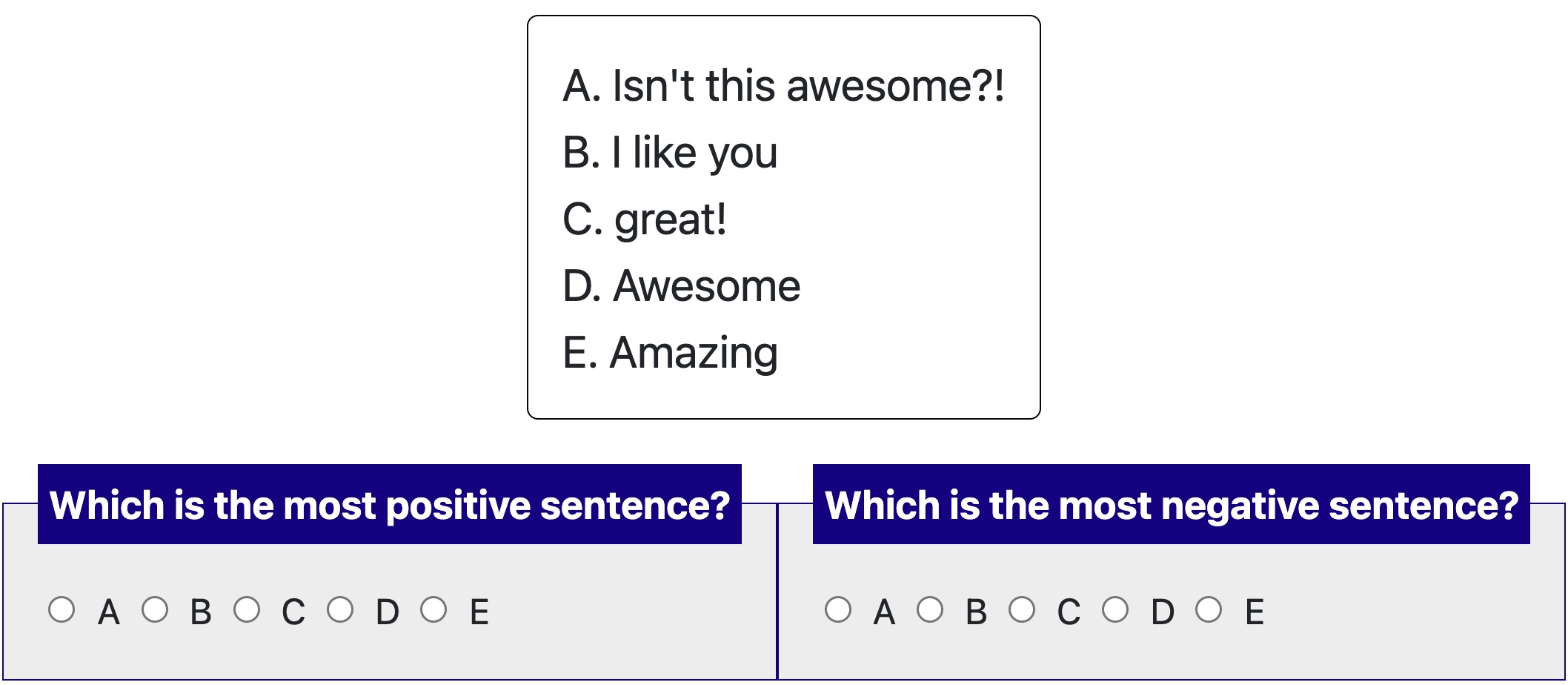}
        \caption{Best-Worst Scaling Annotation}    
        \label{fig:bws}
        
    \end{subfigure}
    \vskip\baselineskip
    %\hfill
    \begin{subfigure}[b]{0.455\textwidth}   
        %\centering 
        \includegraphics[width=\textwidth,left]{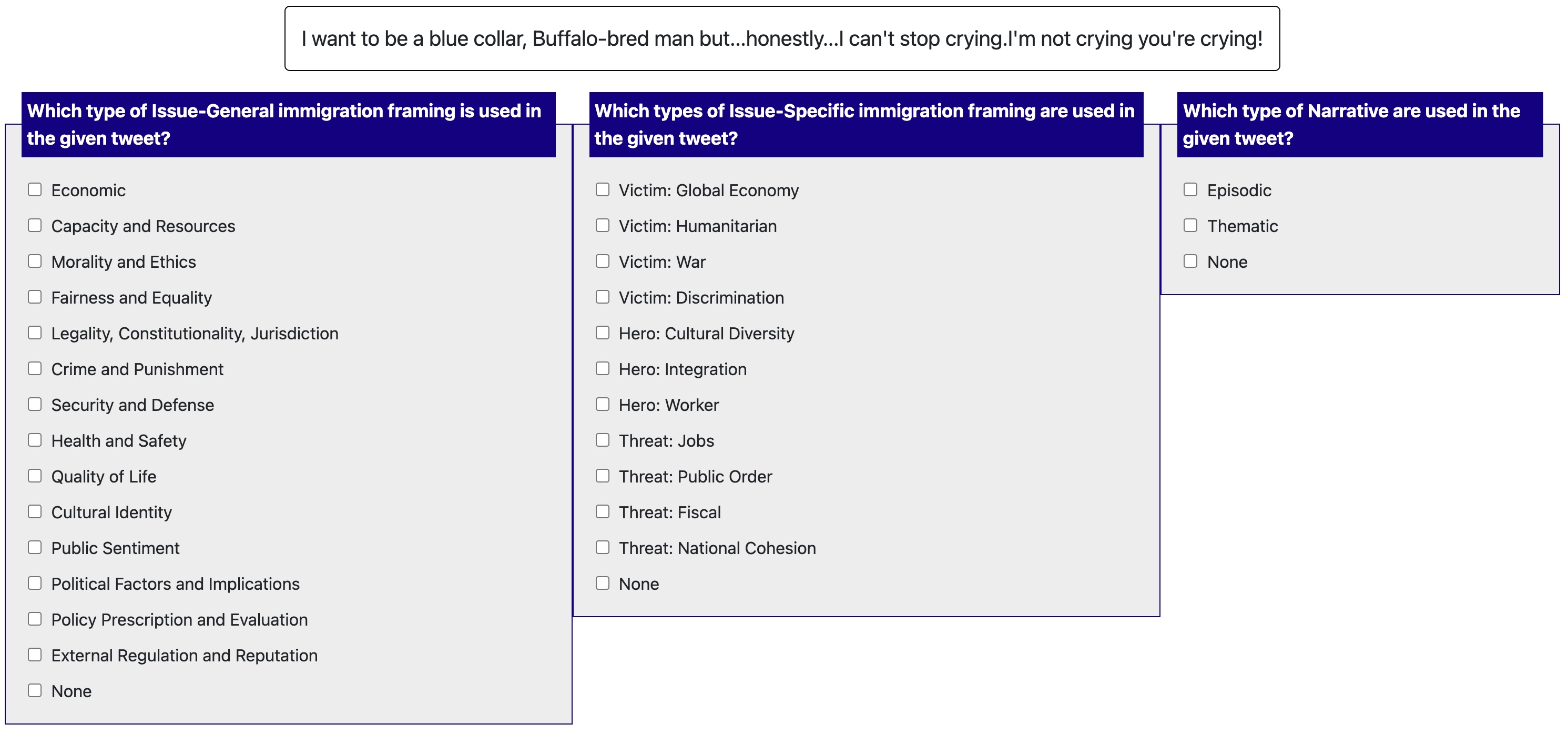}
        \caption{Text Categorization}    
        \label{fig:frames}
    \end{subfigure}
    \hfill
    \begin{subfigure}[b]{0.485\textwidth}   
        %\centering 
        \raggedright
        \includegraphics[width=\textwidth,right]{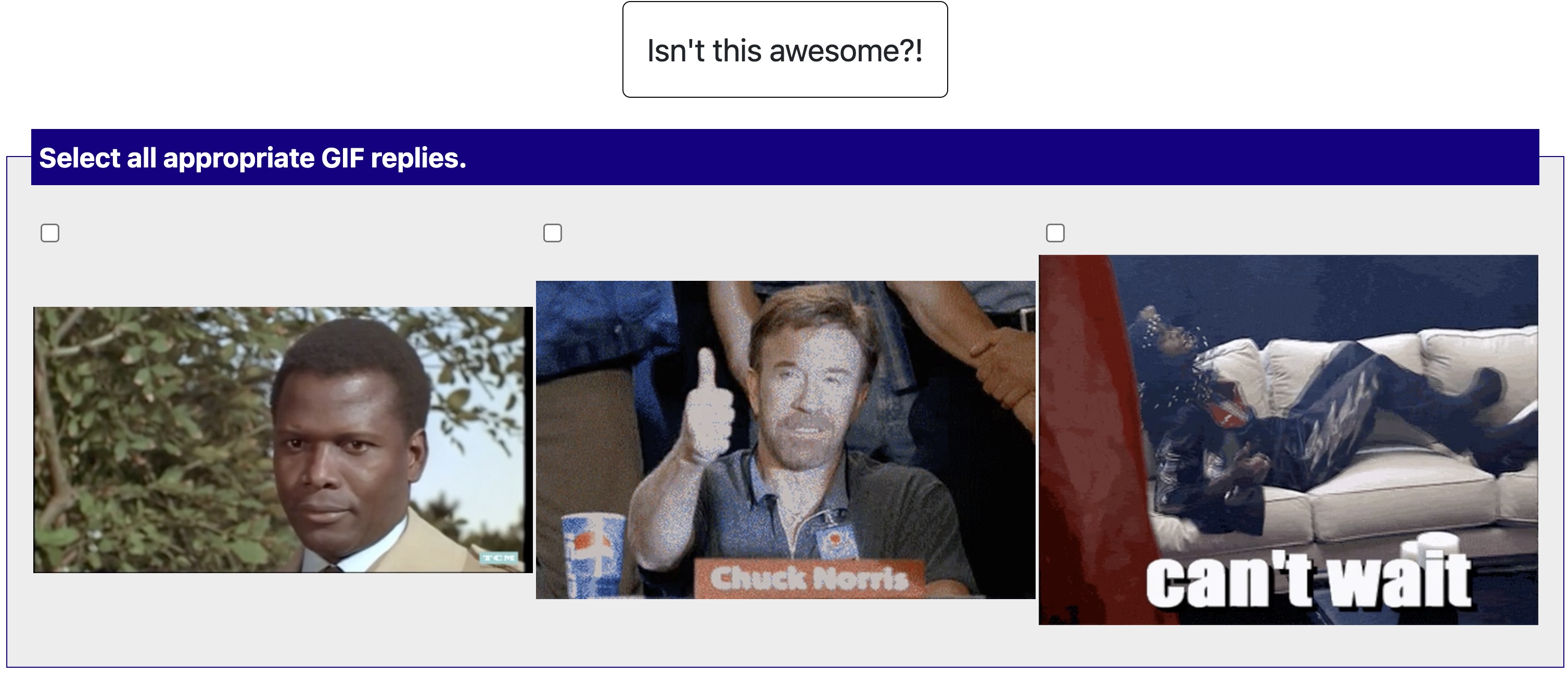}
        \caption[]%
        {{\small Image-based Rating}}    
        \label{fig:image}
    \end{subfigure}
    \vskip\baselineskip
    %\hfill
    \begin{subfigure}[b]{0.455\textwidth}   
        %\centering 
        \includegraphics[width=\textwidth,left]{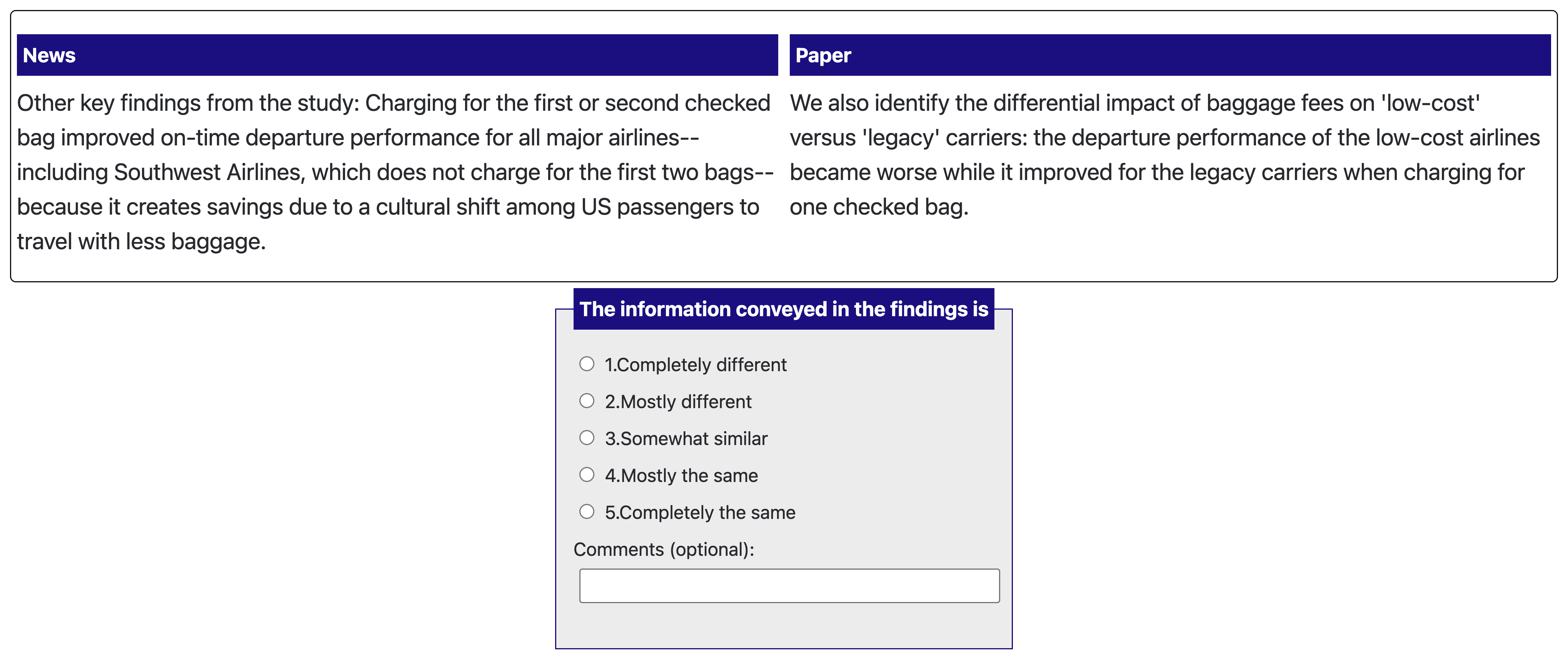}
        \caption{Pairwise comparison}    
        \label{fig:pairwise}
    \end{subfigure}
    \hfill
    \begin{subfigure}[b]{0.485\textwidth}   
        %\centering 
        \raggedright
        \includegraphics[width=\textwidth,right]{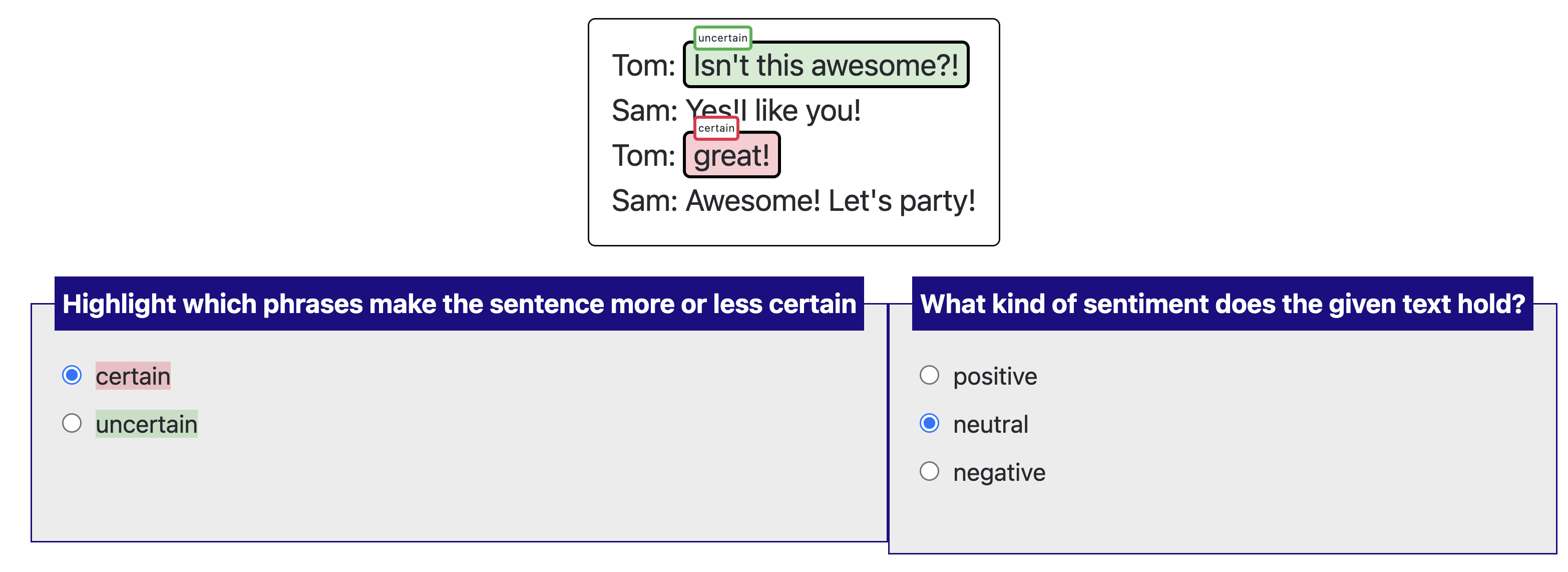}
        \caption[]%
        {{\small Span-based dialogue analysis}}    
        \label{fig:span}
    \end{subfigure}

    \caption{Screenshots of example tasks supported by \potato, which are included as templates. Examples  \ref{fig:likert}-\ref{fig:bws} show single-task annotations, while Example \ref{fig:frames} shows a multitask setup with three multi-select labels. Example \ref{fig:image} shows how \potato supports multimedia as annotation options. \ref{fig:pairwise} shows a pairwise Likert annotation and \ref{fig:span} shows a span-based annotation for dialogue analysis. Examples omit the common interface header that shows annotators how many instances remain and links to the annotation codebook.}
    \label{fig:examples}
\end{figure*}

\section{Deployment and Tasks}
\label{sec:deployment}

\potato is designed with a quickly-deployable Python-based server architecture that can be run locally or hosted on any device. To launch a \potato instance, the deployer first defines a YAML file that specifies the annotation schemes, data sources, server configuration, and any custom visualizations. If \potato is launched without a YAML, the program will provide the deployer the option of following a series of prompts about their task to automatically generate a YAML file for them. A YAML file is then passed to the server on the command line to launch the server for annotation.

Currently, \potato supports eight annotation scheme types: multiple-selection (checkboxes), single-selection (radio buttons), best-worst scaling, Likert scale, free-form text, span-based labels, numbers, and dropdown list. Deployers can easily set up one or more of these schemes in the YAML file---e.g., asking annotators to rate a news article on different dimensions using multiple Likert scales and then summarizing the article in a free text response. For each annotation instance, \potato can take a single document, multiple documents as a list (e.g. dialogue and best-worst-scaling), as well as a dictionary of documents (e.g. a pair of documents for pairwise comparison). \potato will automatically display the instance to annotators based on the input types and the YAML configurations.
The \potato documentation contains example YAML templates for several common annotation tasks such as sentiment analysis, question-answering, and image-based labeling.

\potato is self-hosted and can be served locally or exposed publicly. Each running instance of \potato serves one task, but multiple annotation tasks can be stored in a single \textit{installation}, to be served using different configuration files at different times. \potato allows flexible ways for annotators to login. For internal usage, \potato allows annotators to sign up and log in with email addresses. \potato also allows annotators to directly log in with a URL argument (e.g. username in the crowdsourcing platform), which can be used in crowdsourcing settings where a dedicated link is created for an annotation task. \potato has been tested with popular crowdsourcing platforms including Prolific and Amazon Mechanical Turk.

\potato has been deployed in a variety of annotation settings over a two-year period, including a 27-class annotation scheme for classifying immigration framing \cite{mendelsohn2021modeling}; rating condolences and empathy for Reddit comments with hundreds of words \cite{zhou2020condolences}; best-worst scaling for rating intimacy in questions \cite{pei2020quantifying}; rating Reddit threads for their prosociality \cite{bao2021conversations}; rating, on a Likert-scale, sentences for scientific uncertainty \cite{pei2021measuring}, intimacy in multilingual tweet \citep{pei2022semeval} and similarity in scientific findings \citep{wright2022modeling}; and rating the appropriateness of GIF replies to messages, which showed an animated GIF in the interface \cite{wang2021animated}. 

\fref{fig:examples} shows some of the interfaces from our documentation's example templates. These templates cover a wide range of NLP tasks and can be easily adapted to support a quick start of common annotation tasks. The configuration-based setup of \potato allows researchers to easily share their annotation settings and replicate annotation settings used by existing works. \potato also comes with a dedicated project hub where researchers can easily open-source their annotation project and already includes projects in our previous studies. Such a feature could help to improve the replicability of NLP/ML annotations and we welcome submissions from the entire research community.

\section{Comparison with Existing Systems}

\potato has been developed to fill a key niche left by existing systems for providing visual customization, easy annotator-support features, and rapid development. 
The ultimate goal is to provide simple and comprehensive solutions to common annotation tasks as well as allow personalized design for complex tasks. 
\tref{tab:comparisons} shows the comparisons between \potato and other common text annotation tools over a series of important dimensions including flexibility, productivity, quality, and accessibility. We highlight major differentiators next. Please note that we only compare annotation systems that are currently available for anyone to use, free of cost.

\input{comparison-table}

\paragraph{Flexibility}
\potato is designed to maximize flexibility for a variety of annotation settings. For common annotation tasks like text classification, \potato comes with a wide range of templates and allows a quick start for deployers. However, unlike many existing annotation tools, which provide fixed user interfaces with selected types of annotation tasks (e.g., Doccano offers neither templates nor an editable UI \cite{doccano}), \potato allows deployers to customize their own annotation interface to support diverse needs. For example, \citet{wang2021animated} used animated GIFs as the labels in the annotation and \citet{mendelsohn2021modeling} used a 27-class scheme under three categories, both of which required visual customization to make the task feasible. \potato also allows deployers to easily set up unlimited numbers of similar annotation tasks, which can be especially helpful for multilingual annotations. For example, for all the other data annotation systems, the deployers need to set up separate tasks and guidelines for each language. With \potato, deployers only need to create a sheet containing translated guidelines and \potato's built-in script can help to generate annotation sites for each language. % This function has been used to support the  

\paragraph{Productivity}
\potato is designed to maximize the productivity of both annotators and deployers. While most of the existing annotation tools generally focus on labeling data, \potato supports a series of features that can help with the entire data annotation pipeline. 

\potato allows easily-customizable keyboard shortcuts to allow efficient annotation. For visually or cognitively challenging settings, \potato allows conditional highlights, which helps to reduce task complexity and focus annotators' attention. Finally, active learning can reduce the annotation time needed to curate an informative dataset. Often, annotation tools that offer a highly customizable annotation interface do not also implement productivity features: the open-sourced version of LabelStudio \cite{tkachenkolabel} only supports keypress shortcuts, while Flat \cite{gompel2017folia} supports none of these features. 
For deployers, \potato allows seamless integration with common crowdsourcing platforms like Amazon Mechanical Turk and Prolific. 

\paragraph{Quality control}
Collective high-quality and reliable annotations is the ultimate goal of data labeling tasks and is usually the key to the success of the final ML/NLP systems. \potato comes with a series of quality control feature which helps deployers to reliably collect annotations. While some other annotation systems like Label Studio and WebAnno also support agreement calculation, none of the existing systems come with features that help deployers to improve the annotation quality and analyze factors affecting it. \potato allows deployers to easily set up prestudy qualification tests (annotators have to pass a small test to participate in the full annotation) and attention tests (attention test questions are randomly inserted in the annotation queue as configured by the deployer) to identify unreliable annotators before,  within, and after annotation. \potato also allows deployers to freely insert survey questions before and after the annotation phase. Deployers can easily define different pages of pre- and post-screening questions with minimum effort and \potato also provides a series of templates for common survey questions like user consent and demographic information. Recent studies suggest that the background of annotators has substantial effects on the quality and bias of data labeling and further affects the fairness of ML/NLP models \citep[e.g.,][]{sap2022annotatorsWithAttitudes}. With \potato, researchers can easily collect background information of annotators and analyze the effect of annotator backgrounds on data labeling. 

\paragraph{Accessibility} 
\potato is free to use and actively maintained. While commercial annotation tools like Prodigy \cite{prodigy2017} can come with more functionality, these tools are expensive; for example, Prodigy costs \$390 USD for individual users, and Tagtog \cite{cejuela2014tagtog} costs at least \$59 USD per person per month. These costs are potentially prohibitive for students and researchers without access. \potato is fully open-sourced and is deployed with minimum dependencies. 
%Therefore, unlike other giant annotation systems, \potato is very easy for second-development. 
Moreover, in addition to giving the flexibility to freely define UIs and annotation settings, \potato allows researchers to easily share their annotation settings with a simple YAML file, aiding in replication and future extension of prior work.

\section{Experimental Analysis}
\label{sec:user-study}

\potato was designed to minimize set-up time and per-instance annotation time while maintaining annotation accuracy. Therefore, we conduct a user study to compare the time for setup and annotation time per instance  compared to its free competitors.
We compare \potato's performance on two long and complex tasks, each involving identifying themes and causes in narrative summaries of reports of completed suicides:

\begin{itemize}
    \item \textbf{Task 1} contains long documents, with two annotation schemes with a total of 22 labels.
    The task requires labeling whether each narrative contains each of 13 work-related transitions (e.g., retirement, layoffs) and 9 housing-related transitions. The average document contains 13.4 sentences and 1,180 characters.
    
    \item \textbf{Task 2} is comprised of shorter documents, with the same tasks and labels.
    This alternative version shows only a single sentence of the narrative in Task 1 and asks annotators to judge whether each contains the same categories. The average sentence (annotation item) contains 88 characters.
    
\end{itemize}

\paragraph{Annotation Setup}
One annotator completed 50 annotations for Task 1 and 100 annotations for Task 2 on \potato and a number of freely available and feature-rich annotation tools: Microsoft Excel, Doccano \cite{doccano}, Label Studio \cite{tkachenkolabel}, and LightTag \cite{perry2021lighttag}.
%\footnote{We omitted FLAT \cite{gompel2017folia} after it did not install on the annotator's laptop due to dependency conflicts.} 
The annotator was highly familiar with the task and classification scheme, having annotated 1000$+$ instances of each task prior to this user study, so familiarity with the codebook was not a factor.
Given the complexity of the task, \potato was initialized with 118 keywords to associate with conditional highlights (e.g., retir*, layoff, work*), %\daj{example}
key bindings, and classes included tooltips summarizing each label. 
To test the effect of these productivity-enhancing features, we include a version of \potato that does not include these features, called Inconvenient \potato.
To ensure the same level of familiarity with each document, the documents annotated with each tool are randomly sampled from a larger set of 203,531 documents.

For each annotation tool we measure the time to set up an annotation task without counting time taken to (1) install and familiarize ourselves with the tool (e.g., trial and error in set-up), (2) generate the annotation data files, and (3) write the properties of each label and keywords. 
Each tool was configured as comparably as possible (e.g., keypress shortcuts were always enabled and active learning disabled). 
To reduce the influence of initial unfamiliarity with each tool on per-instance timing, the annotator completed 10 untimed instances. 
Then, for each tool, we record the time spent annotating per instance.

\begin{figure*}[t]
\centering
    \begin{subfigure}[b]{0.3\textwidth}
        \centering
        \includegraphics[width=\textwidth]{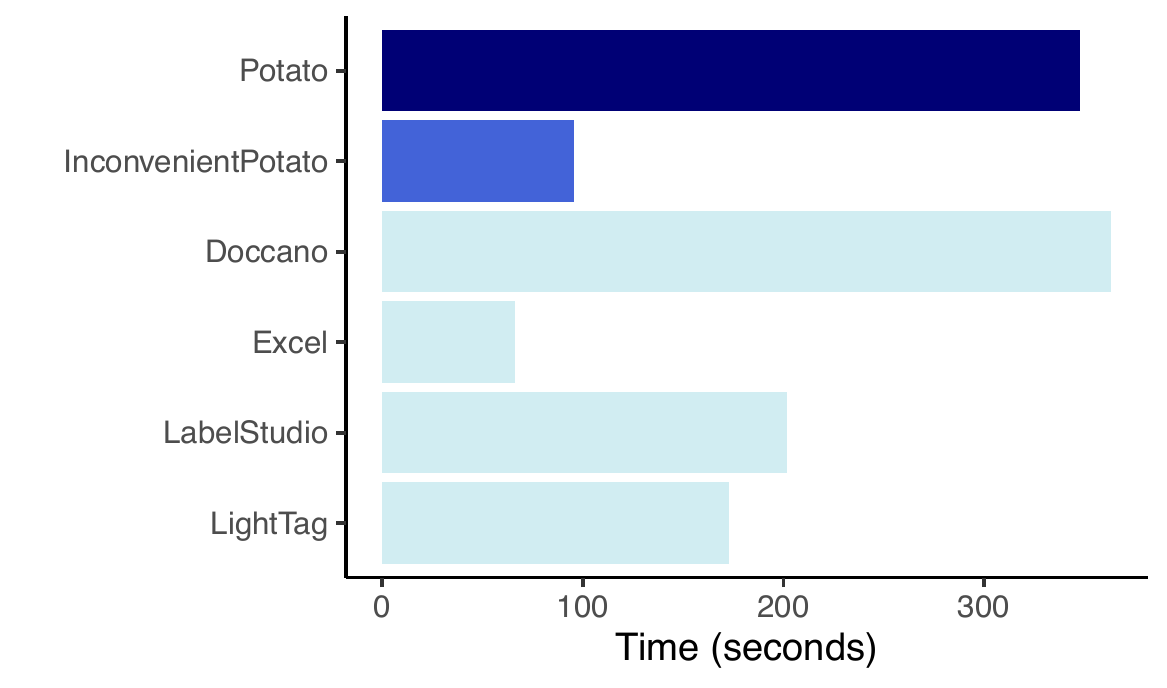}
        \caption{Task Setup Time}
        \label{exp:setup}
    \end{subfigure}
    \hfill
    \begin{subfigure}[b]{0.3\textwidth}
        \centering
        \includegraphics[width=\textwidth]{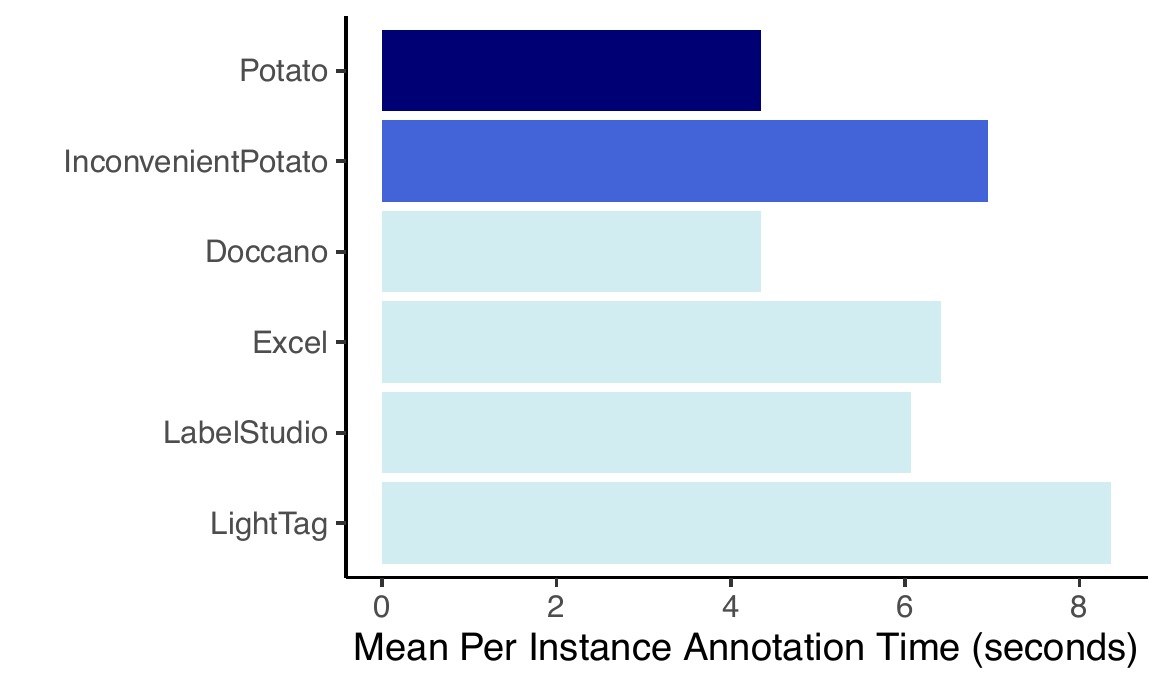}
        \caption{Short Document, 2 tasks, 22 labels}
        \label{exp:short}
    \end{subfigure}
    \hfill
    \begin{subfigure}[b]{0.3\textwidth}
        \centering
        \includegraphics[width=\textwidth]{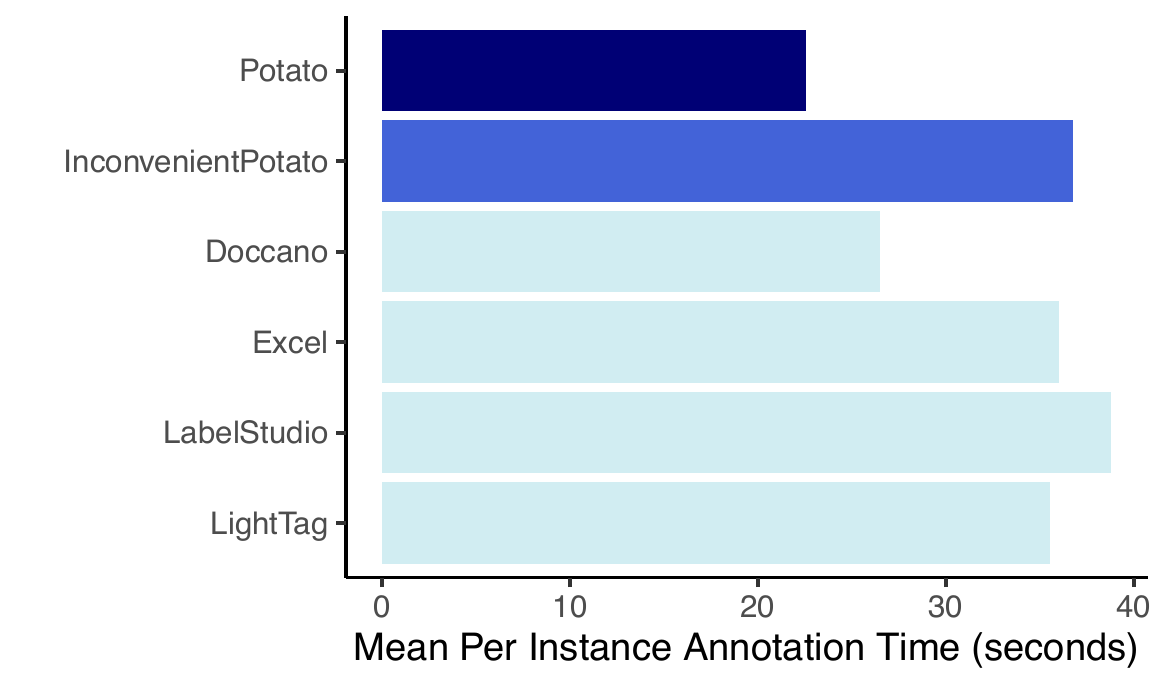}
        \caption{Long Document, 2 tasks, 22 labels}
        \label{exp:long}
    \end{subfigure}    
    \caption{Times from our user study (\sref{sec:user-study}) to (a) set up a task; (b) annotate one short document or (c) annotate one long document show the time savings of \potato.}
    \label{fig:experiment}
\end{figure*}

\paragraph{Results}
Across both annotation tasks, \potato is approximately 30-50\% faster than competitors like Excel, Label Studio, and LightTag (Figure~\ref{fig:experiment}b-c).
Annotating short documents is comparable in time to Doccano (\ref{fig:experiment}b), while long documents are slightly faster in \potato (\ref{fig:experiment}c). 
Without convenience features like conditional highlighting, key mappings, and tooltips, \potato's per-instance annotation time increases to be more comparable with other tools. We conclude that the difference in per-instance annotation time is likely attributable to these design features.

Including convenience features increases task setup time (Figure~\ref{fig:experiment}a), taking just over 4 extra minutes to configure at set-up. Our results suggest that, compared to using \potato with convenience features, the base setup without convenience features has lower overall task time (including setup) until an annotator has seen $\sim$20 long documents or $\sim$100 short documents.   %\aka{Actual numbers: 18 and 97, but I'm rounding since my error bars are so giant!} 
%\potato is also relatively quick and easy to set up. 
We note that \potato without convenience features takes less time to set up than Label Studio and LightTag; with these three features, \potato takes roughly the same amount of time to set up as Doccano, even though Doccano only supports one feature (keypress shortcut). 

The two tasks we chose share features with many common NLP annotation tasks that make them well-suited to a system like \potato. We highlight two comparative observations across the interface from the user study.
First, Doccano and \potato have the most annotator-friendly interfaces, which allow for fast coding. For instance, custom keypress shortcuts allow us to create 22 different bindings that make logical sense to the annotator. Unlike other tools, these two tools did not require any use of the mouse (e.g., others required pressing submit with the mouse), which reduced the annotation time; in short document tasks, where the time to read the document is low, these time savings become especially important. The default page layout in \potato also better supports content interpretation; for example, the text and labels frequently fit on one page with no scrolling, and by placing the text on top of the labels, the annotator did not need to scroll down in order to read the text---and since the annotator had the keypress bindings memorized, the task could often be accomplished with no scrolling. Finally, since other tools often required uploading data to external servers, there was often a load time of 1-2 seconds per document; again, saving this time with locally-deployed tools was especially salient with short documents.

Second, both tasks involve assigning a large number of independent labels. The keyword highlights allow the annotator to quickly identify which subset of these labels are likely to be relevant to the document, while the key mappings allow them to quick apply the correct labels. Keyword highlights are particularly useful for longer documents (e.g., Task 1), because they help identify the text that is likely relevant to a given label, which is often buried in a large amount of irrelevant text (e.g., most labels apply to one of $\sim$13 sentences). Additionally, if none of the labels applied to a document, the annotator needed to ensure that she did not overlook a relevant phrase in the long text; in \potato the lack of relevant keywords allowed us to quickly confirm that none of the labels applied, while without keyword highlights, the annotator read these documents twice. 
%\daj{what is ``this'' referring to here:} 
The time savings associated with keyword highlights likely explains the slight per-instance advantage \potato has over Doccano for long documents but not short documents.

\section{Conclusion and Future Plans}

\potato distinguishes itself with a comprehensive suite of productivity-enhancing features that allow annotators to efficiently and accurately label data and researchers to quickly configure complex tasks on a wide range of data types. \potato was created both for the computational scholar and the overburdened student or crowdworker, looking to annotate more data in their limited time.

\potato has been in continuous development for over two years and will continue to be developed to support new task designs, easier management, and faster annotation. 

For management, we aim to have (1) a unified GUI for deployers to create new tasks and manage existing tasks, (2) a GUI that supports real-time monitoring of annotation process, mirroring tools like Webanno \cite{yimam2013webanno}, and (3) integration with common social media platforms to display content with original interfaces (e.g. displaying tweets with their native UI).

For annotators, we aim to support simple linguistic search to help annotators find and prioritize instances to annotate, and to support personalization in aspects such as annotators' visualization and keybindings. We also plan to conduct experiments and explore different design choices to reduce annotator burn-out.

\section*{Acknowledgments}

The authors thank members of the Blalablab and all the annotators who have used \potato throughout the years for their feedback on how to make \potato better.  This material is based in part upon work supported by the National Science Foundation under Grant No IIS-1850221.

\section{Ethics and Broader Impacts}

As a highly configurable annotation tool, \potato's biggest ethical and societal implications will likely come from the questions the tool is used to answer and the ways in which researchers choose to deploy the tool. 
\potato was built with accessibility, social responsibility, and usefulness at the forefront, and the tool's default settings afford a range of values-driven practices, which we will discuss below. However, a major risk is that \potato requires researchers to self-regulate when encouraging researchers to opt into ethical values often proves unsuccessful \cite{hagendorff2020ethics}. For instance, the tool does not build in any safeguards against unethical questions or harmful applications \cite{buolamwini2018gender,mitchell2019model,benjamin2019race} and does not actively prevent the exploitation of crowdworkers \cite{irani2013turkopticon,shmueli2021beyond}. Moreover, since \potato is a tool designed to improve the efficiency of typical prediction task workflows, it cannot address existential critiques of machine learning (e.g., harms of classification as a practice).

\potato was created using principles of universal design, prioritizing broadly experienced ease of use, low effort, intuitiveness, flexibility, tolerance for error, and perceptibility of key information \cite{persson2015universal}. Since \potato is uniquely annotator-focused, rather than deployer-focused, tasks are readily designed in a way that maximizes worker wellbeing and productivity. The application's design is largely accessible and inclusive and the tool contains many of the types of features that crowdworkers find useful (e.g., low effort to configure, reduces cognitive burden of complex tasks, easy to correct errors by going back, login flow that supports screen readers and doesn't use captcha, annotation guidelines readily visible in tooltip and hyperlink) \cite{zyskowski2015accessible,swaminathan2017crowd}. That said, certain features of the interface may be inaccessible for workers. Moreover, tools like \potato can worsen annotators' mental health by promoting fragmented work, multitasking, and poor work-life balance \cite{williams2019perpetual} and by displaying triggering text without masks or warnings \cite{shmueli2021beyond}. Many of these potential accessibility and psychological harms can be addressed through improvements in the interface. Because of the ease of secondary development --- especially adding new HTML front-end templates --- \potato allows the research community to explore more design opportunities for inclusive annotation and responsible crowdsourcing. Ideally, a future version of this tool would use community-led design to develop more universally accessible, inclusive templates for users \cite{spiel2020nothing}.

In designing \potato, we prioritized developing mechanisms for just, equitable compensation. By allowing annotators to track time spent on the task, \potato facilitates paying crowdworkers a fair hourly wage rather than the per-task payment schemes that frequently lead to low hourly wages \cite{fort2011amazon,gray2019ghost}. A key accessibility feature, the timer promotes flexibility (e.g., allows people to take longer or build in microbreaks) instead of imposing needlessly restrictive per-task time requirements that can create barriers for disabled workers \cite{zyskowski2015accessible}. Our goal in creating \potato was to empower and support the annotator. For instance, although we piloted a timer to alert annotators when the expected task time had elapsed, we ultimately removed this feature in order to eliminate additional stress.

Another important problem in computational social research is inaccurately labeled and biased datasets, which are a cause of inequitably felt downstream harms \cite{olteanu2019social,blodgett2020language,mehrabi2021survey}. \potato may have the potential to reduce many common sources of bias by promoting high-quality annotations: convenience features lower cognitive load and reduce reliance on personal heuristics that may increase bias; researchers can use tooltips to provide specific, easily accessed instructions to minimize anticipated sources of bias; since the base annotation time is lower, and there are no per-instance annotation time limits, annotators may feel less pressure to label faster at the expense of poor annotation quality. However, \potato may amplify the researchers' own biases in the data: annotators may rely too heavily on keyword highlights and tooltips, which can bias the data if keywords common in minority communities are over- or underrepresented, or the tooltip text does not include instructions relevant to certain communities in the data. Future experiments can study the effect of \potato's productivity-enhancing features on mitigating or amplifying different types of bias. 

Finally, an important goal in developing \potato was to facilitate studying complex social questions without being limited by existing labeled data: since the tool makes it easier and faster to design complex tasks and collect data, researchers can think critically about what problems would be most beneficial and impactful, and design tasks that actually answer those questions \cite{wiens2019no,abebe2020roles}. Since \potato facilitates the deployment of multilingual tasks, researchers can more easily test the generalizability of their results across linguistic and cultural contexts \cite{joshi2020state}. A major challenge in applied machine learning is the lack of diversity among researchers \cite{orife2020masakhane}; since \potato is free, open-sourced, and easy to use, we hope the tool will facilitate participation by scholars who are not associated with well-funded R1 universities and also by community members outside academia. 

\bibliography{references}
\bibliographystyle{acl_natbib}

\appendix

\end{document}

%% file: comparison-table.tex
\begin{table*}[t!]
\newcommand{\tabincell}[2]{\begin{tabular}{@{}#1@{}}#2\end{tabular}}
\newcolumntype{P}[1]{>{\centering\arraybackslash}p{#1}}
\centering

\resizebox{\textwidth}{!}{
   %\rowcolors{5}{}{lightgray}
   
   % TODO Fix the centering and alignment with https://texblog.org/2019/06/03/control-the-width-of-table-columns-tabular-in-latex/
    \begin{tabular}{|c | r | c | c | c | c | c | c | c | c | c | c |}
 %\multicolumn{1}{c}{} & \multicolumn{7}{c}{\textbf{Flexibility}} & \multicolumn{4}{c}{\textbf{Productivity}} & \multicolumn{3}{c}{\textbf{Surveyflow}} & \multicolumn{1}{c}{\textbf{Privacy}} & \multicolumn{2}{c}{\textbf{Price and Availability}} 
 \cline{3-12}

\multicolumn{2}{c|}{} & Label Studio & Doccano & FLAT & LightTag & Prodigy & Tagtog & FITAnnotator & BRAT & WebAnno/INCEpTION & \potato
 
\\ \hline \hline
 
%                                 LS       DO       FL       LT       PR       TT       FI       BR      WA         PO

 & Multiple Schema                & \cmark &        &        & \cmark &        & \cmark &        &  \cmark      &  \cmark  & \cmark \\ \cline{2-12}
 & Multimodal                  & \cmark & \cmark &        & \cmark & \cmark &        &        &        &        & \cmark \\ \cline{2-12}
 & Span-Based Annotation       &   \cmark &   \cmark    &        & \cmark  & \cmark   &    \cmark    &        & \cmark &   \cmark &  \cmark    \\ \cline{2-12}
 %& Multilingual \& Multi-domain&        &        &        &        &        &        &        &        &        & \cmark \\ \cline{2-12}
%Many Labels                 & \cmark & \cmark & \cmark & \cmark &        & \cmark & \cmark &        &        & \cmark \\ \hline
%Templates                   & \cmark &        &        & \cmark & \cmark & \cmark &        &        &        & \cmark \\ \hline
\multirow{-4}{*}{Flexibility} &        Editable UI                 & \cmark &        & \cmark &        &  \cmark &        & \cmark &        &        & \cmark \\ \hline  \hline
\rowcolor{lightgray}
 &        Active Learning   &   \cmark*     &        &        &        & \cmark &        & \cmark &        &  \cmark  & \cmark \\ \cline{2-12} 
\rowcolor{lightgray}
&        Conditional Highlighting    &        &        &        &        &        &        &        &        &        & \cmark \\ \cline{2-12}
\rowcolor{lightgray}
\multirow{-3}{*}{Productivity} &        Keyboard Shortcuts           & \cmark & \cmark &        & \cmark & \cmark &        &        &   \cmark     &        & \cmark \\ %\cline{2-12}
\hline \hline
%\rowcolor{lightgray}
%&        Compatibility w/ Crowd Work &        &        &        &        &        &        &        &        &        & \cmark \\ \cline{2-12}
%\rowcolor{lightgray}
%\multirow{-5}{*}{Productivity} &        Automatic Task Assignment   &        &        &        &        &        &        &        &        &        & \cmark \\ \hline \hline 

\multirow{4}{*}{Quality control}  & Prestudy Qualification Test &        &        &        &        &        &        &        &        &        & \cmark \\ \cline{2-12}
%&   Agreement tracking & \cmark* &        &        &        &        &        &        &        &  \cmark  & \cmark \\ \cline{2-12}
&        Attention Test &        &        &        &        &        &        &        &        &        & \cmark \\ \cline{2-12}
&        Behavioral Tracking &        &        &        &        &        &        &        &        &        & \cmark \\ \cline{2-12}
&  Pre- and Post-screeing Questions &        &        &        &        &        &        &        &        &        & \cmark \\  \hline \hline
%&        Runs Locally                & \cmark & \cmark & \cmark &        & \cmark & \cmark & \cmark & \cmark & \cmark & \cmark \\ \hline \hline 
\rowcolor{lightgray}
&        Open-Source                 & \cmark & \cmark & \cmark &        &        &        &        &  \cmark    &   \cmark     & \cmark \\ \cline{2-12}
\rowcolor{lightgray}
&        Easy Sharing and Replicating   &  & & &        &        &        &        &     &       & \cmark \\ \cline{2-12}
\rowcolor{lightgray}
\multirow{-3}{*}{Accessibility} &        Price                       & Free   & Free   & Free   & Free for academia & \$390 & \$59/person/month & Not available & Free & Free  & Free \\ \hline

    \end{tabular}}
    \caption{Comparisons between \potato and other text annotation systems over four themes. * means the feature is not available for the free plan. %Potato provides free text annotation solutions with flexibility for diverse needs, a series of productivity features as well as privacy-reserving designs. 
    % \daj{to-do: fix the vertical bars disappearing due to the new cmark macro centering.}
    %\jp{add compatibility with AMT and prolific, behavioral tracking}
    }
    \label{tab:comparisons}
\end{table*}